\documentclass{article}
\usepackage{graphicx}
\date{}
\usepackage{amsmath}
\usepackage{amssymb}
\usepackage[backend=biber,style=numeric-comp]{biblatex}
\usepackage{listings}
\usepackage[frozencache=true,cachedir=.]{minted} 

\usepackage{xcolor}

\usepackage[most]{tcolorbox}
\usepackage{xparse}
\usepackage{newfloat}
\usepackage{caption}
\usepackage{enumitem}

\usepackage[bookmarks=false]{hyperref}
\usepackage{orcidlink}

\usepackage{soul}

\lstdefinestyle{promptstyle}{
  basicstyle=\ttfamily\footnotesize,
  columns=fullflexible,
  breaklines=true,
  frame=single,
  framerule=0.4pt,
  rulecolor=\color{black},
  showstringspaces=false,
  keepspaces=true
}

\lstdefinestyle{mypython}{
  language=Python,
  basicstyle=\ttfamily\small,
  numbers=left,
  numberstyle=\tiny,
  stepnumber=1,
  numbersep=8pt,
  xleftmargin=0.24in,
  frame=none,
  showstringspaces=false,
  breaklines=true,
  breakatwhitespace=true,
  tabsize=2,
  keepspaces=true,
  columns=fullflexible,
  keywordstyle=\color{blue!60!black},
  commentstyle=\color{green!40!black},
  stringstyle=\color{orange!60!black},
}

\title{Automatic Design of Optimization Test Problems with Large Language Models}
\author{%
Wojciech~Achtelik\,\orcidlink{0009-0001-7363-4448}\thanks{This work was partially supported by the National Science Centre, Poland, grant number 2023/49/B/ST6/01404.}\thanks{Wojciech Achtelik, Hubert Guzowski, and Maciej Smo{\l}ka are with the Faculty of Computer Science, AGH University of Krakow, Al. A. Mickiewicza 30, 30-059 Kraków, Poland (e-mail: \{wachtelik,guzowski,smolka\}@agh.edu.pl).}%
\and
Hubert~Guzowski\,\orcidlink{0000-0002-9678-0384}\footnotemark[2]%
\and
Maciej~Smo{\l}ka\,\orcidlink{0000-0002-3386-0555}\footnotemark[2]%
\and
Jacek~Ma{\'n}dziuk\,\orcidlink{0000-0003-0947-028X}\thanks{Jacek Ma{\'n}dziuk is with the Faculty of Mathematics and Information Science, Warsaw University of Technology, ul. Koszykowa 75, 00-662 Warsaw, Poland and with the Faculty of Computer Science, AGH University of Krakow, Al. A. Mickiewicza 30, 30-059 Krakow, Poland (email: mandziuk@mini.pw.edu.pl).}%
}

\begin{document}

\maketitle

\begin{abstract}
The development of black-box optimization algorithms depends on the availability of benchmark suites that are both diverse and representative of real-world problem landscapes. Widely used collections such as BBOB and CEC remain dominated by hand-crafted synthetic functions and provide limited coverage of the high-dimensional space of Exploratory Landscape Analysis (ELA) features, which in turn biases evaluation and hinders training of meta-black-box optimizers. We introduce Evolution of Test Functions (EoTF), a framework that automatically generates continuous optimization test functions whose landscapes match a specified target ELA feature vector. EoTF adapts LLM-driven evolutionary search, originally proposed for heuristic discovery, to evolve interpretable, self-contained numpy implementations of objective functions by minimizing the distance between sampled ELA features of generated candidates and a target profile. In experiments on 24 noiseless BBOB functions and a contamination-mitigating suite of 24 MA-BBOB hybrid functions, EoTF reliably produces non-trivial functions with closely matching ELA characteristics and preserves optimizer performance rankings under fixed evaluation budgets, supporting their validity as surrogate benchmarks. While a baseline neural-network-based generator achieves higher accuracy in 2D, EoTF substantially outperforms it in 3D and exhibits stable solution quality as dimensionality increases, highlighting favorable scalability. Overall, EoTF offers a practical route to scalable, portable, and interpretable benchmark generation targeted to desired landscape properties.
\end{abstract}

\section{Introduction}

The development and analysis of black-box optimization algorithms heavily rely on the availability of diverse and challenging benchmark functions. These test problems represent the principal means by which algorithmic performance is evaluated, different solvers are compared, and their respective strengths and weaknesses are understood. Ideal benchmarks should be computationally efficient, interpretable, and representative of the complex characteristics found in real-world optimization tasks. However, widely-used benchmark suites like the Black-Box Optimization Benchmarking (BBOB)~\cite{COCO} and IEEE CEC Competition series \cite{cec} rely on artificial functions. These often do not reflect the complexities of real-world applications, such as Hyperparameter Optimization (HPO) \cite{hpo_x_ela}, and neural network training \cite{malan_nn_benchmarking}. Moreover existing benchmarks lack diversity needed for training Meta-Black-Box Optimization algorithms \cite{wang2025}.

To systematically characterize and compare optimization problems, the field of Evolutionary Computation (EC) has broadly adopted Exploratory Landscape Analysis \cite{ela}. ELA provides a set of quantitative metrics describing the fitness landscape, which have proven valuable for tasks such as automated algorithm selection \cite{Preuss_ELA_ASP, Tanabe_ELA_ASP}. Despite the utility of ELA, a significant deficiency persists: existing benchmark suites do not adequately cover the high-dimensional space of possible ELA feature vectors \cite{malan_nn_benchmarking}. This coverage gap implies that algorithms are often evaluated on a limited and potentially biased set of problem landscapes, which complicates the generalization of performance findings to new, unobserved problems. While new functions can be designed manually to address these gaps, the process is intricate, time-intensive, and demands considerable mathematical expertise.

In response, the research community has investigated methods for the automated generation of benchmark functions. Prominent approaches include affine transformations \cite{bench-MA-BBOB}, Genetic Programming (GP) \cite{gp_evolution, wang2025, munoz-space-filling, evolving-problems} and Neural Networks (NNs) \cite{foga_nn}. The above methods, however, exhibit significant limitations. Affine transformations span only a subset of the entire ELA feature space and remain fundamentally constrained by the structural properties of the underlying BBOB functions~\cite{wang2025}. GP can be constrained by a predefined grammar of operators and tends to produce convoluted symbolic expressions that are difficult to analyze. NNs, while highly flexible, are opaque models themselves, yielding a set of network weights rather than a transparent mathematical formulation. A critical, shared weakness of GP and NN-based methods is the limited control over the analytical properties of the resulting functions.

\subsection{Contribution}
This paper introduces a new method for automatic test function design that leverages the capabilities of modern Large Language Models (LLMs). LLMs present a unique combination of attributes well-suited for this task. Having been trained on extensive corpora of scientific and mathematical literature, they possess a substantial "expert prior" relevant to optimization theory and practice \cite{llamea, llm-sr}. This enables them to generate human-readable, symbolic functions in code, offering a compelling alternative that balances the structural rigidity of GP with the flexibility of NNs. By producing interpretable expressions, LLMs facilitate a more profound understanding of the generated problem landscapes and shift the role of the human expert from direct design to high-level guidance and validation.

The primary contribution of this work is the introduction and validation of ELA-guided benchmark generation using LLMs. We demonstrate the ability of LLMs to generate novel, non-trivial optimization problems that conform to a target ELA feature vector. 

Through a series of experiments, we demonstrate that the LLM-generated functions not only match the target ELA characteristics but also induce similar performance rankings among a portfolio of optimization algorithms, indicating their efficacy as analytical surrogates. The proposed LLM-based methodology represents a promising, efficient, and accessible approach to creating diverse and insightful benchmarks required to advance the field of Black-Box Optimization.

\section{Background and Motivation}

The automated generation of mathematical functions and algorithms, traditionally addressed by methods like GP~\cite{gp_evolution}, has been recently revolutionized by the advent of LLMs. Unlike GP, which is constrained by a predefined grammar, LLMs leverage vast training corpora of text and code to generate syntactically correct and semantically meaningful expressions in a more flexible, human-readable format. This capability has established LLMs as powerful tools for code generation and scientific discovery.

A pioneering application of LLMs in the considered domain is FunSearch \cite{funsearch}, which successfully discovered new mathematical functions to solve challenging problems in extremal combinatorics. Building on this concept of evolutionary improvement, subsequent research has applied LLMs to the automatic design of optimization algorithms themselves. Notable examples include Algorithm Evolution using Large Language Model (AEL) \cite{ael}, Evolution of Heuristics (EoH) \cite{eoh} and LlaMEA \cite{llamea}, which maintain a population of optimizer implementations and use an LLM as a mutation operator to evolve better-performing heuristics. In a task closely related to our objective, LLMs have also been successfully employed for Symbolic Regression \cite{symbolic-regression}, where models like LLM-SR \cite{llm-sr} find concise mathematical formulas to fit data, capitalizing on the "expert prior" gained from scientific literature. Another example is LLaMEA-BO, which leverages LLMs to generate novel acquisition functions for Bayesian Optimization \cite{li2025llameabolargelanguagemodel}. 

Prager et al. in \cite{foga_nn} propose a pipeline for constructing synthetic test functions whose ELA descriptors match a user-specified target. Their method first generates a random point cloud $X$ in the decision space together with random objective values $y\in[0,1]$, and then optimizes only the objective values $y$ so as to minimize the Euclidean distance between the resulting ELA feature vector and a prescribed target vector. Once an optimized point cloud $(X, y^*)$ is obtained, they fit a neural network surrogate that deliberately overfits this dataset, and the trained network is used as a new benchmark function.

We consider the problem they formulate highly relevant: it directly targets controlled diversity in benchmark suites, described via landscape properties such as ELA features. However, their approach exhibits a fundamental scalability bottleneck. The optimization variable is the full vector $y$, whose dimensionality equals the point-cloud size; with their recommended sampling rate of $250\cdot D$, the resulting optimization problem scales linearly in $D$ but quickly becomes extremely high-dimensional.

This motivated us to explore alternative approaches that sample functions  directly, without an explicit intermediate dataset construction step. In this context, we view LLMs as implicit distributions over executable program text (e.g., Python code). Sampling from such a distribution yields candidate test functions, potentially avoiding the expensive optimization over $y$ while still enabling diversity control.

\subsection{Research Gap}
The above works collectively demonstrate that LLMs excel at tasks involving the discovery and evolution of code, functions, and algorithms. However, to the best of our knowledge, their application to the targeted generation of benchmark functions for Black-Box Optimization, guided by specific landscape properties, remains unexplored. While EoH and LlaMEA evolve optimizers, we evolve test functions to rigorously evaluate those optimizers.

\section{Proposed method}

We propose Evolution of Test Functions (EoTF), a framework that extends EoH~\cite{eoh} from algorithm design to test function generation. While EoH evolves optimization algorithms, EoTF applies the same LLM-driven evolutionary architecture to evolve interpretable Python functions whose fitness landscapes match a target ELA feature vector.

\subsection{Problem Formulation}

Extending the methodology introduced in \cite{foga_nn}, we frame test function generation as a landscape feature-matching problem. Given a target vector of ELA features $\boldsymbol{\phi}^* \in \mathbb{R}^k$, the goal is to construct a single-objective continuous function $f: \mathcal{X} \rightarrow \mathbb{R}$, with $\mathcal{X} \subset \mathbb{R}^d$, whose landscape features approximate $\boldsymbol{\phi}^*$. In contrast to prior approach that use NNs to model landscapes~\cite{foga_nn}, yielding uninterpretable weight configurations, we search over a space $\mathcal{F}$ of human-readable symbolic expressions. Formally, we seek a function $f^*$ minimizing:
\begin{equation}
    f^* = \arg\min_{f \in \mathcal{F}} \|\boldsymbol{\phi}_{f} - \boldsymbol{\phi}^*\|_2
    \label{eq:ela_minimization}
\end{equation}
where $\boldsymbol{\phi}_{f}$ denotes the feature vector extracted from a sample of the landscape induced by $f$. Regarding feature selection, we utilize the set of ELA features identified in \cite{foga_nn} and extracted via the \texttt{pflacco} \cite{pflacco} package. This selection is motivated primarily by the need for direct comparability with the prior NN-based generation method. Furthermore, this specific subset provides a balanced representation of landscape characteristics. The features include:
\begin{itemize}
    \item \texttt{ela\_meta.lin\_simple.adj\_r2}: Adjusted coefficient of determination for a linear regression model without interaction terms~\cite{ela}.
    \item \texttt{ela\_meta.lin\_w\_interact.adj\_r2}: Adjusted coefficient of determination for a linear regression model including interaction terms~\cite{ela}.
    \item \texttt{ela\_meta.quad\_simple.adj\_r2}: Adjusted coefficient of determination for a quadratic regression model without interaction terms~\cite{ela}.
    \item \texttt{ela\_meta.quad\_w\_interact.adj\_r2}: Adjusted coefficient of determination for a quadratic regression model including interaction terms~\cite{ela}.
    \item \texttt{ela\_distr.skewness}: Skewness of the sample's objective values~\cite{ela}.
    \item \texttt{nbc.nb.fitness.cor}: Correlation between the fitness values and the in-degree of points in the nearest-better graph~\cite{detecting_funnel}.
    \item \texttt{nbc.nn\_nb.sd\_ratio}: Ratio of the standard deviations of all nearest neighbor distances to the nearest better neighbor distances~\cite{detecting_funnel}.
    \item \texttt{fitness\_distance.fitness\_std}: Standard deviation of the objective values within the sample~\cite{fitness_distance}.
\end{itemize}
It is important to note that our method is not structurally sensitive to this specific choice of features. The method is flexible and can be adapted to arbitrary subsets of ELA features. LLMs support context windows exceeding 1M tokens \cite{team2024gemini}, scaling the method to include a more comprehensive or distinct set of landscape descriptors is  feasible and remains a promising direction for future study.

In our experiments, target feature vector $\boldsymbol{\phi}^*$ is calculated as an average of ELA features calculated for 100 different independent samples, each generated with a different random seed, from the same target benchmark problem instance. All feature vectors are normalized via min-max scaling, where the minimum and maximum values for each feature are obtained from sampling across all problems within a given benchmark suite and dimensionality.

\subsection{Evolution of Test Functions}
We adopt the operator taxonomy introduced in EoH~\cite{eoh}, comprising one initialization strategy, two exploration operators, and three mutation operators. While the operator semantics remain unchanged, we design domain-specific prompts tailored to the test function generation task. Full prompt templates are provided in Appendix~\ref{app:prompts}.

\begin{figure}
\includegraphics[width=\columnwidth]{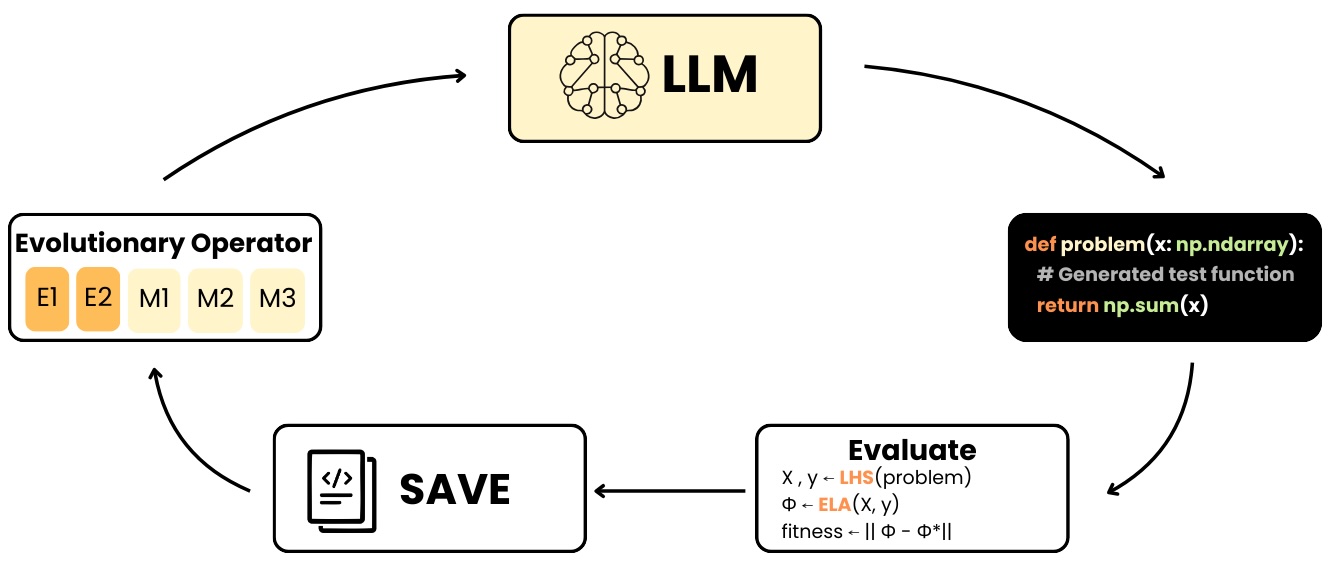}
\caption{A summary of the proposed EoTF method.}
\end{figure}

Compared to generating complete optimizers, producing benchmark functions is structurally simpler: the output is a Python function without the object-oriented complexity. However, the task imposes its own challenges: the LLM must synthesize mathematically coherent, landscape-diverse expressions that exhibit prescribed statistical and structural properties. An example of a generated function is shown in Listing~\ref{lst:generated_function}. The function combines quadratic, trigonometric, and cubic terms to produce a non-trivial fitness landscape. This particular instance was produced by Gemini~2.0~Flash to match the ELA profile of a reference function from the BBOB suite. 

\begin{listing}[ht]
\begin{minted}[linenos,xleftmargin=0.24in]{python}
import numpy as np

def problem(x: np.ndarray) -> float:
 quadratic_term = 0.13 * (
   x[0] ** 2 + x[1] ** 2
 )
 cosine_modulation = 0.13 * np.cos(
   x[0] - x[1]
 )
 linear_interaction_term = 0.045 * (
   x[0] + x[1] + x[0] * x[1]
 )
 skewed_cubic_term = 0.027 * (
   x[0] ** 3 + 0.5 * x[1] ** 3
 )
 bias = 0.05
 return (
   quadratic_term
   + cosine_modulation
   + linear_interaction_term
   + skewed_cubic_term
   + bias
 )
\end{minted}
\caption{Example Python function generated by the EoTF method.}
\label{lst:generated_function}
\end{listing}

\subsection{Interpretability and Reusability Advantages}

A distinctive advantage of employing LLMs for benchmark function generation lies in the nature of the final artifacts. The output of our method is a self-contained Python function, minimal in dependencies, lightweight in execution, and readily portable. Such functions can be shared without additional complexity associated with distributing benchmarks based on NNs or other machine learning models, which often require large model files, specialized frameworks, and non-trivial installation procedures. The reduced dependency footprint facilitates reproducibility and adoption in the research community.

Equally important is the interpretability of the generated functions. While the decision-making process of the LLM during generation remains opaque, the produced code is fully transparent. Researchers can inspect the source, parse docstrings, and directly trace the mathematical structure of the function. This property enables both empirical and theoretical analysis: one can study gradient behavior, curvature, modality, or other analytical characteristics directly from the symbolic formulation. In contrast, NN-based benchmarks only yield weight sets, making such analyses far more cumbersome.

We argue that this interpretability significantly enhances the practical utility of LLM-generated functions. Beyond their immediate use as benchmarks, they serve as a foundation for theoretical investigations into problem landscape design. Future work may leverage this property to systematically categorize LLM-generated functions, identify recurring structural motifs, and streamline the creation of new test functions tailored to specific algorithmic evaluation needs.

\subsection{Scalability to Higher Dimensions}
Practical black-box optimization tasks frequently operate in high-dimensional search spaces. Scaling to such spaces represents a critical bottleneck for prior generative methods. For instance, the recent NN-based approach~\cite{foga_nn} struggles to generalize across dimensions: reported results for three-dimensional functions are notably worse than those for two-dimensional cases. Additionally, NN methods are typically constrained to fixed input sizes. In contrast, LLMs offer an advantage by generating symbolic Python code. By using dimension-agnostic operations (e.g., via \texttt{numpy} arrays \cite{harris2020array}), the generated functions can be dimension-independent. A single symbolic expression produced by the LLM can thus define a valid landscape for arbitrary input dimensions without modification or retraining.

\subsection{Model selection}

LLMs are evolving at a rapid pace, with new models emerging every few months. For our evaluation, we select both the lightweight models and the more advanced ones capable of reasoning. We adopt Gemini 2.0 Flash~\cite{comanici2025gemini25pushingfrontier} as the default model for experiments, motivated by its use in AlphaEvolve \cite{alphaevolve}. In addition, we include Gemini 2.5 Flash, and Gemini 3.0 Flash in our experiments to analyze the effect of using more recent and more capable LLMs. According to LMArena \cite{lmarena} (as of 21 December 2025), these models rank 89th, 50th, and 3rd, respectively, on the global leaderboard, indicating substantial diversity in capability. These differences are also reflected in latency and cost per token trade-offs.

\subsection{Function Selection}
In our setup, we employ multiple target test functions to assess the quality of LLM-generated functions. The primary source of target ELA features is the BBOB benchmark set \cite{COCO}. We execute EoTF on each function independently. A notable challenge in this setting is data contamination, as many research papers analyzing the ELA features of BBOB functions are likely included in the LLMs’ training corpora. To mitigate this risk, we supplement the benchmark with selected functions from MA-BBOB and randomly generated ELA values. Consistent with \cite{foga_nn}, we focus on 2, and 3-dimensional problem instances. A key distinction from \cite{foga_nn} is that our approach is substantially faster and can be readily applied to high-dimensional domains.

\section{Experimental Setup} 
\label{sec:experimental_setup}

Our experiments are designed to address the following four research questions regarding the feasibility and efficacy of LLM-driven test function generation.
\begin{enumerate}
    \item RQ1: Can EoTF effectively generate functions with target ELA features?
    \item RQ2: Does the method maintain its effectiveness when applied to problems beyond canonical benchmarks?
    \item RQ3: What is the impact of LLM on generated test function quality?
    \item RQ4: How does the method scale with number of dimensions?
\end{enumerate}

\subsection{RQ1: EoTF Performance}

Our main baseline is the NN-based method~\cite{foga_nn}. Furthermore, to justify the selection of the LLM-based evolution architecture; we compare EoTF against LlaMEA, another LLM-based method. LlaMEA maintains a population of candidate solutions, apply mutation and selection operators implemented via LLM prompts. Since LlaMEA focuses on generation of metaheuristics for continuous optimization, we adopt it to the domain of benchmark function generation by redefining the prompts and the fitness function as the distance between a candidate's ELA vector and a predefined target vector. We also introduce a stateless Zero Shot baseline, where all generations are independent and utilize the same prompt. This baseline tests whether EoTF's added complexity is necessary, or whether a simpler prompt-only LLM approach can achieve comparable quality. For consistency, EoTF, LlaMEA, and the Zero Shot method all employ \textbf{Gemini 2.0 Flash} with a fixed budget of 250 queries per method. This model was chosen for its balance of performance and efficiency, as observed in preliminary tests and its use in AlphaEvolve \cite{alphaevolve}. We employ the standard BBOB suite \cite{COCO}, utilizing the first instance of all 24 functions. 

\subsubsection*{Evaluation Metrics}
The quality of the synthesized functions ($f^*$) is assessed via a comprehensive evaluation protocol:

\begin{itemize}
    \item \textbf{Method Comparison.} To ensure that the generated functions are robust and not artifacts of a specific sampling seed, we do not rely on fitness function values (best distance). Instead, we recalculate the ELA vector over 100 independent samples (of the same generated function) and calculate a median distance for each function. Then we use these values to compare different methods and report a single number: percentage of wins between methods. Method A is better than method B for a given problem (function id) if function generated using method A has lower median distance of 100 samples than method B. 

    \item \textbf{Visual Verification.} For 2D functions, we provide a qualitative assessment using scatter plots to visually compare the landscape of generated test functions against their respective targets.

    \item \textbf{Optimizer Ranking.} To validate that the generated functions present similar optimization challenges to the targets, we compare the ranking of diverse optimization algorithms on function sets. Following the methodology of \cite{foga_nn}, we employ a fixed-budget setup ($10,000 \cdot$ $D$ evaluations) across a portfolio of seven algorithms from Nevergrad \cite{nevergrad}: Differential Evolution, PSO, EMNA, Nelder-Mead, dCMA-ES, Cobyla, and Random Search. We aggregate results using Critical Difference diagrams, derived from a non-parametric Friedman test with Nemenyi post-hoc analysis from \texttt{autorank} \cite{autorank}.
\end{itemize}

\subsection{RQ2: EoTF Generality}

We employ the standard BBOB suite \cite{COCO}, utilizing the first instance of all 24 noiseless functions. While these provide a recognized baseline, there is a risk that modern LLMs have memorized the source code for these specific problems during pre-training. To mitigate the risk of data contamination and test generalization, we constructed a custom suite of 24 "hybrid" functions using the MA-BBOB generation methodology. We generated 24 hybrid functions by pairing the BBOB functions in a cyclic ring topology (Function 1 with 2, 2 with 3, \dots, 24 with 1) and setting the mixing coefficient $\alpha = 0.5$. This forces the LLM to generate code for novel ELA features, that are unlikely to exist in its training data.

\subsection{RQ3: LLM selection impact}

This question investigates whether more advanced LLMs, with superior reasoning abilities, produce higher-quality benchmark functions. We compare three models from the Gemini family, representing different tiers of capability, cost, and latency: \textbf{Gemini 2.0 Flash}, \textbf{Gemini 2.5 Flash}, and \textbf{Gemini 3.0 Flash}. The primary metric for this experiment is the average sampled ELA distance achieved for each model. The dataset used is a standard BBOB. Our hypothesis is that more capable models will converge to solutions with a lower distance to the target ELA vector, indicating a better fit to the desired landscape properties.

\subsection{RQ4: EoTF Scalability}

The results presented in~\cite{foga_nn} do not scale well in terms of both latency and quality of solutions. Especially concerning is a discrepancy between best distances and sampled distances for $3D$ problems and the fact that the authors restricted the reported results to $2D$ and $3D$ only, due to  computational complexity of their method. Our method scales much better with number of dimensions. We present the results for $D=2,3,4,$ and $5$. The goal is to prove that quality of solutions does not degrade with dimensionality. We report average median sampled distance of all functions in the BBOB.

\section{Results}
This section presents empirical results answering the four research questions posed in Section~\ref{sec:experimental_setup}.

\subsection{RQ1: EoTF Performance}

\begin{figure}[ht!]
\centering
\includegraphics[width=0.8\columnwidth]{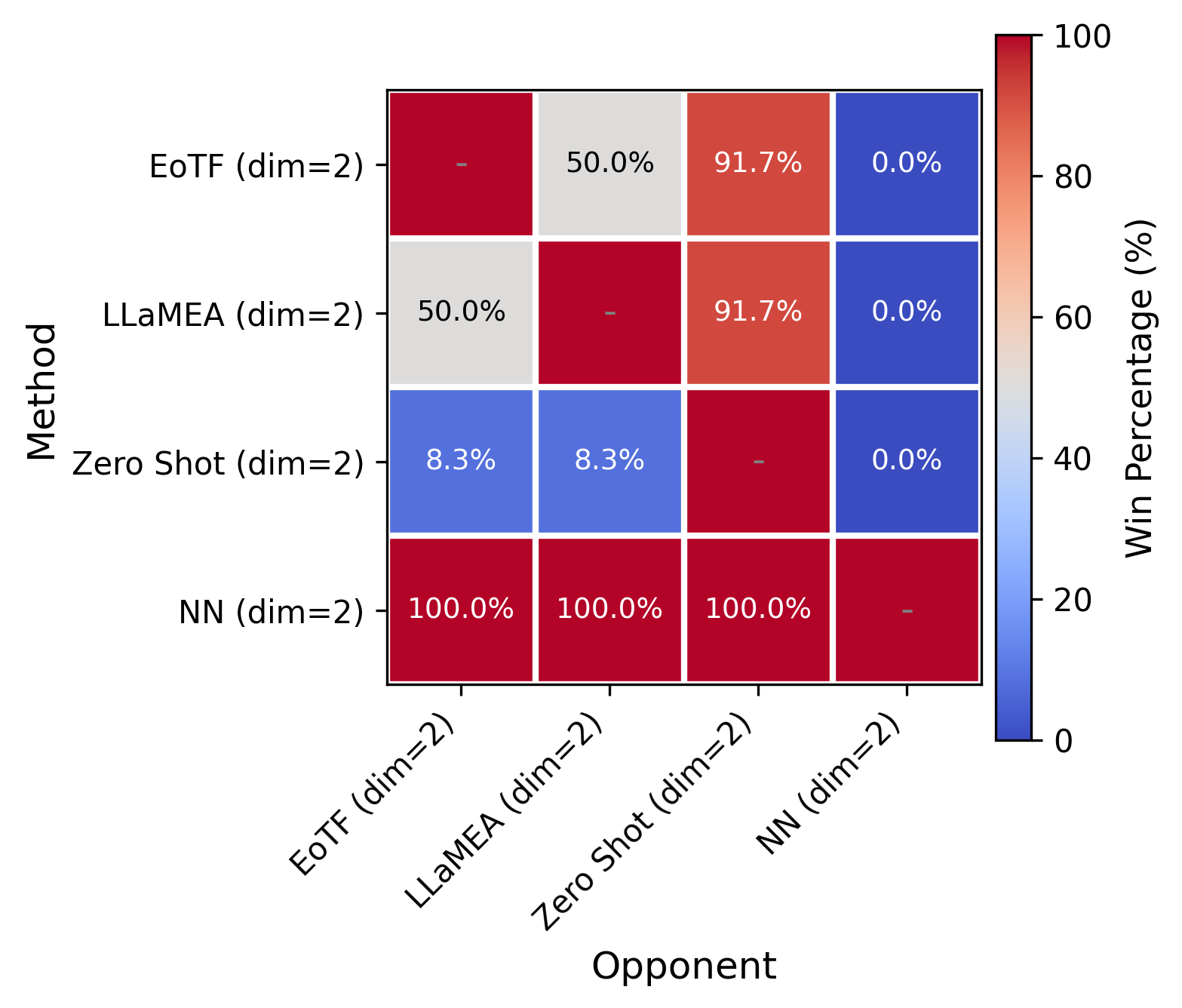}
\caption{Win-percentage matrix comparing EoTF with three baselines: NNs \cite{foga_nn}, zero shot, and LLaMEA.}
\label{fig:method_comparison_2d}
\end{figure}

\begin{figure}[ht!]
\centering
\includegraphics[width=0.8\columnwidth]{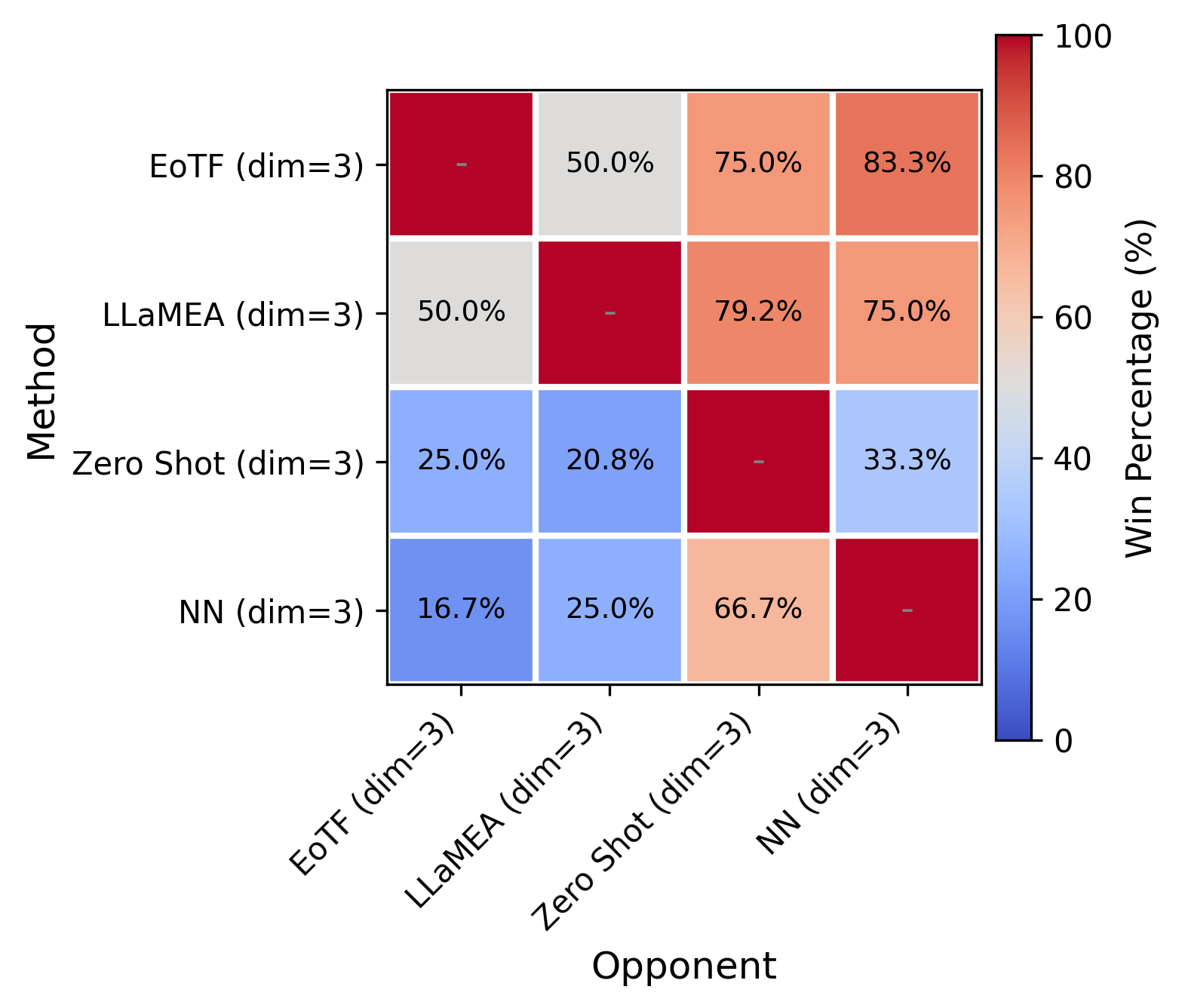}
\caption{Win-percentage matrix comparing EoTF with three baselines for $D = 3$.}
\label{fig:method_comparison_3d}
\end{figure}

To provide a direct statistical comparison, we compute pairwise win rates between all methods. For each BBOB function, the method achieving a lower median ELA distance (computed over 100 samples) is declared the winner.  Figure~\ref{fig:method_comparison_2d} presents results for two dimensional problems. The win percentage represents the fraction of the 24 BBOB functions where a row method outperforms a column method. As can be seen in the figure, neural networks are more accurate than LLMs and win on all 24 problems. Since practical optimization problems have usually dimensionality higher than 2, we also compare methods for 3D search space. Figure~\ref{fig:method_comparison_3d} presents the results, which are drastically different -- now NNs lose in more than 75\% of the cases.

The results generally show that EoTF performs well for $D=3$, however, its advantage over LLaMEA is slight; while EoTF has a higher win rate against the baseline, the two methods perform similarly in direct comparison. This similarity can be explained by similar architectures of EoTF and LLaMEA  and also similar prompts used. These findings suggest that the specific choice of evolutionary architecture may be secondary in this context. This is a point that should be studied in more detail in future work. Simultaneously, both matrices indicate that the Zero Shot approach fails to match the performance of EoTF and LLaMEA. In 2D, the LLM-based evolutionary approaches outperform the Zero Shot baseline in over 90\% of the cases, while in 3D, this advantage decreases slightly to 75\%.

\begin{figure}[ht!]
\centering
\includegraphics[width=\columnwidth]{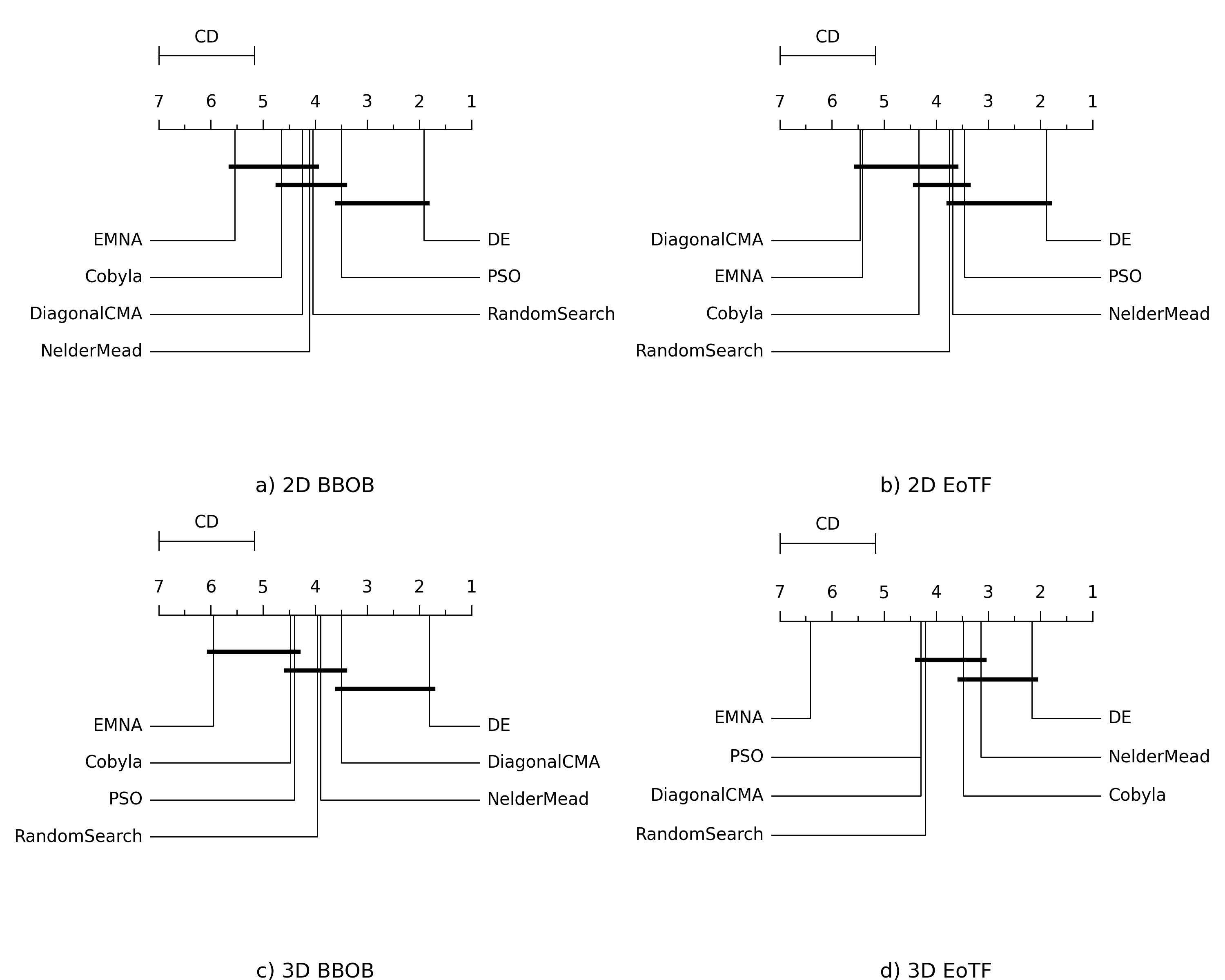}
\caption{Critical distance plot comparing results of a portfolio of algorithms on original and generated (EoTF) problems.}
\label{fig:algorithm_benchmark_comparison}
\end{figure}

Figure \ref{fig:algorithm_benchmark_comparison} presents critical difference plots comparing algorithm rankings on the original BBOB benchmark suite against the EoTF-generated functions for 2D and 3D problem instances. In the 2D setting, the optimizer ordering is almost unchanged and deviates from BBOB in only two aspects. First, Nelder-Mead and Random Search swap positions; this is a minor effect given that their average ranks are very close and fall within overlapping CD groups. Second, DiagonalCMA deteriorates substantially: while it is mid-ranked on BBOB, it becomes the worst performer on the EoTF-generated set. This systematic drop suggests that the synthesized 2D landscapes may be relatively less favorable to CMA-style adaptation, potentially increasing the challenge for covariance-based search dynamics. While this warrants deeper analysis (e.g., ELA-feature shifts that particularly affect CMA behavior), the remainder of the ranking structure is preserved, indicating that EoTF retains the overall optimizer difficulty profile with only localized deviations.

Similarly, for 3D the ordering pattern is also stable, with the main change being an effective swap between Cobyla and DiagonalCMA: DiagonalCMA performs worse on EoTF than on BBOB, whereas Cobyla improves markedly and moves into the better-performing group. The ordering of the remaining optimizers is largely retained, reinforcing that EoTF preserves the relative performance of most methods when moving to higher dimension. An additional noteworthy difference is that EMNA becomes more clearly separated from the rest on the EoTF-generated set: its rank remains worst, but the gap to the mid-performing cluster increases, suggesting that the generated 3D functions accentuate the mismatch between EMNA’s modeling assumptions and the induced landscapes without altering the overall rank hierarchy.

\begin{figure}[ht!]
\centering
\includegraphics[width=0.8 \columnwidth]{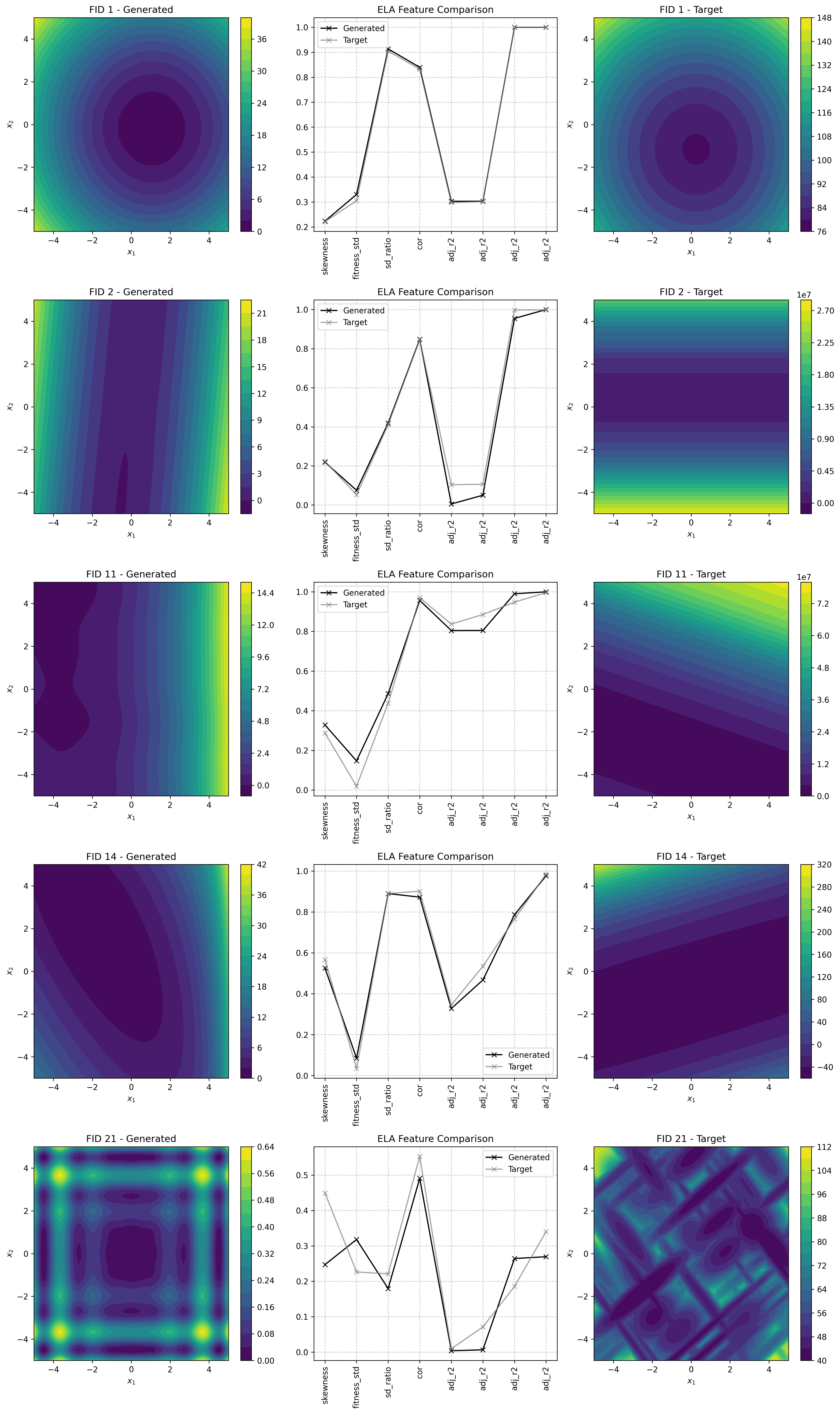}
\caption{Contour grids for few selected functions generated to match BBOB problems with function ids: 1, 2, 11, 14, 21. On the left are the generated functions, on the right original problems. ELA feature values are presented in the middle column}
\label{fig:contour_grids}
\end{figure}

Figure \ref{fig:contour_grids} depicts contour visualizations comparing the fitness landscapes of BBOB benchmark functions with their corresponding EoTF-generated counterparts in 2D. Visual inspection reveals that for the first four function instances (FID 1, 2, 11, and 14), the generated landscapes exhibit strong qualitative similarity to their target functions, successfully capturing essential  characteristics such as the gradual increase in fitness values toward the domain boundaries and the preservation of global structure. However, a notable discrepancy is observed for FID 21, where the generated function fails to reproduce the complex, rugged multimodal landscape. The generated variant exhibits a considerably simplified structure with clearly defined global features. This limitation is quantitatively reflected in the ELA feature space, where FID 21 demonstrates a substantially larger feature distance compared to the other instances. 

When contrasted with the neural network-based function generation approach proposed by \cite{foga_nn}, our method produces landscapes that more closely approximate the original BBOB functions. The NN-generated problems exhibit artifacts consistent with overfitting, resulting in landscapes that appear less smooth and lack the natural structure inherent to the benchmark suite.

\subsection{RQ2: EoTF Generality}
A critical concern in LLM-driven code generation is the potential for data contamination, specifically, whether the model is solving the problem analytically or simply recalling memorized solutions for known BBOB functions. To address this, we compared the framework's performance on the standard BBOB suite against the novel MA-BBOB suite.

\begin{figure}[ht!]
\centering
\includegraphics[width=\columnwidth]{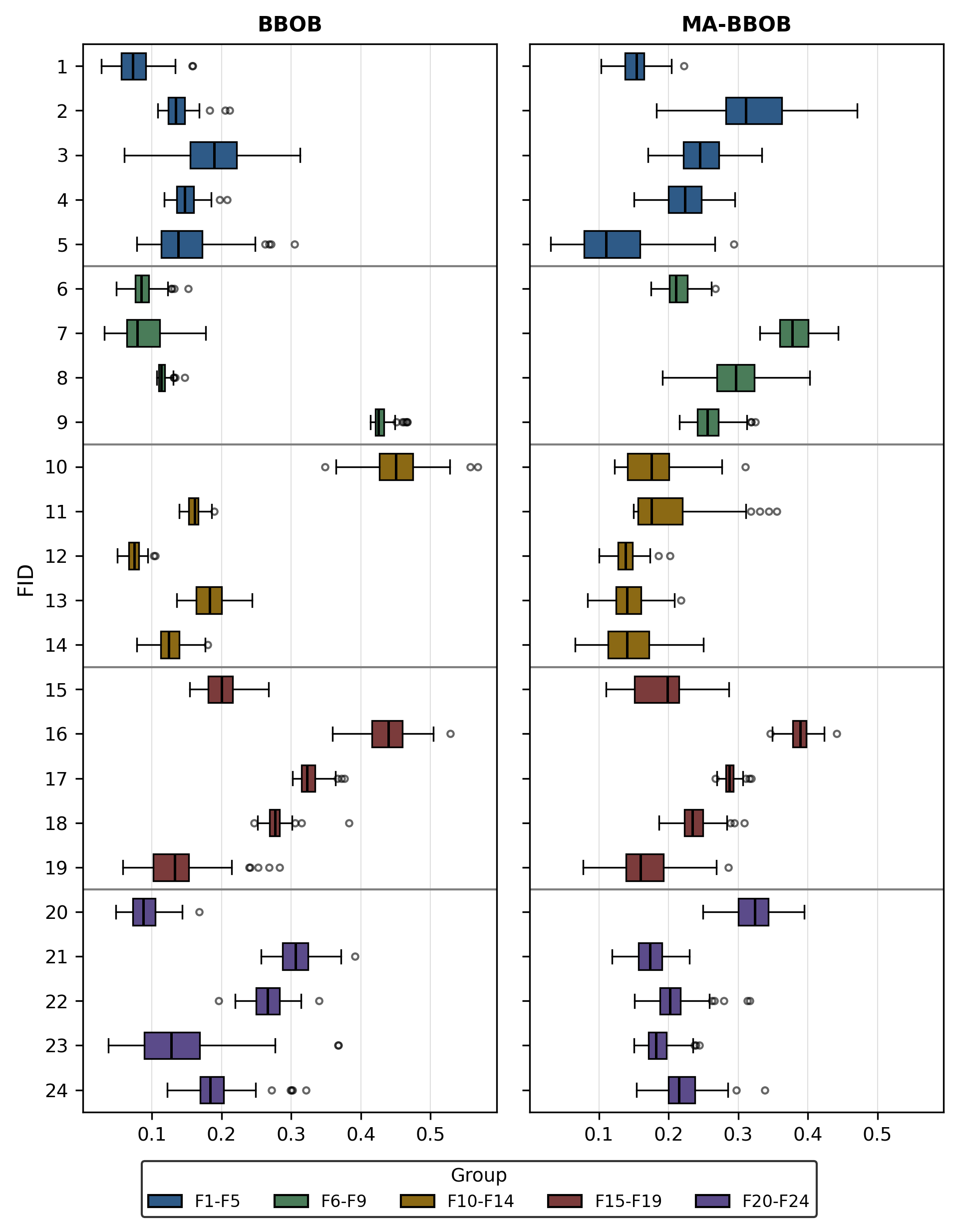}
\caption{Comparison of the sampled distances for BBOB and MA-BBOB problems.}
\label{fig:comparison_ma_bbob}
\end{figure}

Figure \ref{fig:comparison_ma_bbob} presents a side-by-side comparison of the final ELA distances achieved for the 24 standard functions versus the 24 hybrid functions. As observed, the distribution of distances is remarkably similar between the two sets. In several instances, the framework achieved a better fit for the novel MA-BBOB functions than for the standard BBOB functions, and vice versa. While this is not a direct one-to-one comparison, as the underlying  target functions and their landscapes differ significantly, the consistency in performance across both suites is a strong indicator of robustness. It demonstrates that the method generalizes effectively to unseen problem instances and does not rely solely on the memorization of standard benchmark code. The results shown are for 2D functions. Overall MA-BBOB has better median distance values for around 42\% of functions.

To complement the MA-BBOB generalization study with a qualitative check, we manually inspected the generated code for all 24 BBOB targets in the $D=2$ setting. Only one generated function ($f=10$) contained an explicit reference to the Ackley function; the remaining functions were implemented directly using elementary mathematical primitives (e.g., $\sin$, $\cos$, $\sqrt{\cdot}$, $|\cdot|$) and did not match the structure of standard BBOB reference implementations. While manual inspection cannot rule out all forms of implicit memorization, it provides evidence that the framework is not simply recalling benchmark code.

\subsection{RQ3: LLM selection impact}

\begin{figure}[ht!]
    \centering
    \includegraphics[width=0.8\columnwidth]{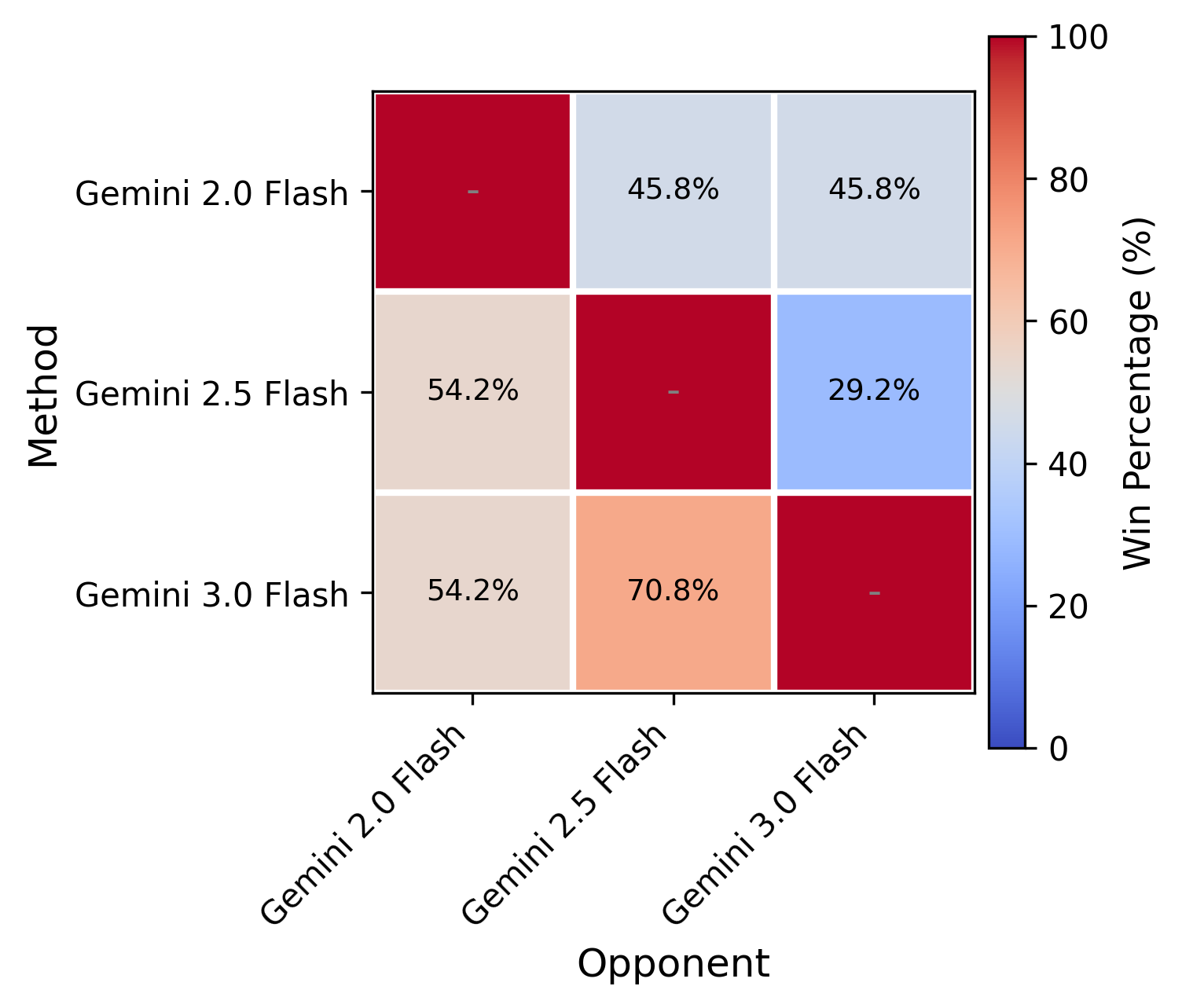}
    \caption{Win-percentage matrix comparing three Gemini model variants across 24 two-dimensional BBOB functions. Each cell indicates the percentage of functions for which the row method achieved a lower median ELA distance than the column method.}
    \label{fig:matrix_comparison_llm}
\end{figure}

To evaluate the influence of model capability on function generation quality, we conducted pairwise comparisons across all 24 benchmark functions. The win-percentage matrix (Figure~\ref{fig:matrix_comparison_llm}) presents the proportion of functions for which the row method achieved a lower median ELA distance than the column method. This asymmetric matrix reveals a clear capability hierarchy among the evaluated models.

\textbf{Gemini 3.0 Flash} demonstrates superior performance, achieving wins against \textbf{Gemini 2.5 Flash} on 70.8\% of functions and against \textbf{Gemini 2.0 Flash} on 54.2\% of functions. The intermediate model, \textbf{Gemini 2.5 Flash}, outperforms \textbf{Gemini 2.0 Flash} on 54.2\% of functions while losing to \textbf{Gemini 3.0 Flash} on the majority of comparisons. These results support the hypothesis that increased model capability correlates with improved function generation quality, as measured by ELA distance to the target. In particular, \textbf{Gemini 2.0 Flash} remains competitive despite being the least capable model, winning approximately 45\% of pairwise comparisons against both successors. This finding underscores the viability of the EoTF approach even with earlier-generation LLMs.

\begin{figure*}[ht!]
    \centering
    \includegraphics[width=\textwidth]{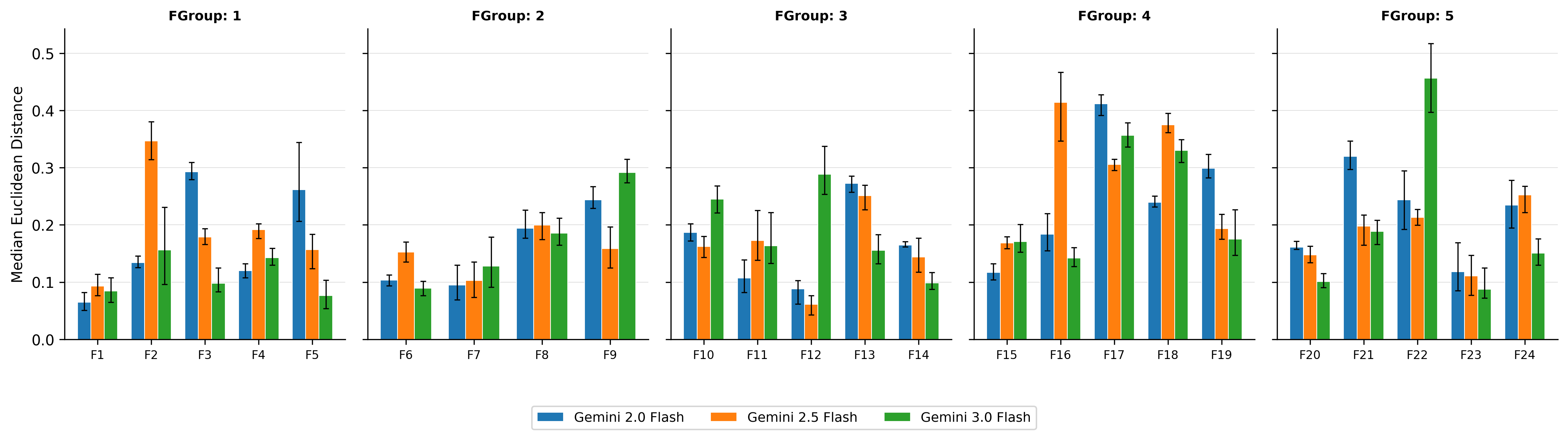}
    \caption{Median sampled ELA distances for all 24 two-dimensional BBOB functions, grouped by function class. Error bars represent the range between $q_{25}$ and $q_{75}$.}
    \label{fig:llm_comparison}
\end{figure*}

Figure~\ref{fig:llm_comparison} provides a function-level breakdown of median sampled distances, with error bars indicating the interquartile range ($q_{25}$--$q_{75}$). While the win-percentage matrix establishes an overall ranking, the per-function analysis reveals substantial overlap in performance distributions across models. The relatively modest absolute differences suggest that model capability improvements yield incremental rather than transformative gains in generation quality for this task. Nevertheless, as LLM capabilities continue to advance alongside declining inference costs, we anticipate cumulative improvements that will further enhance the practical applicability of the proposed method.

\subsection{RQ4: EoTF Scalability}
\begin{figure}[ht!]
\centering
\includegraphics[width=\columnwidth]{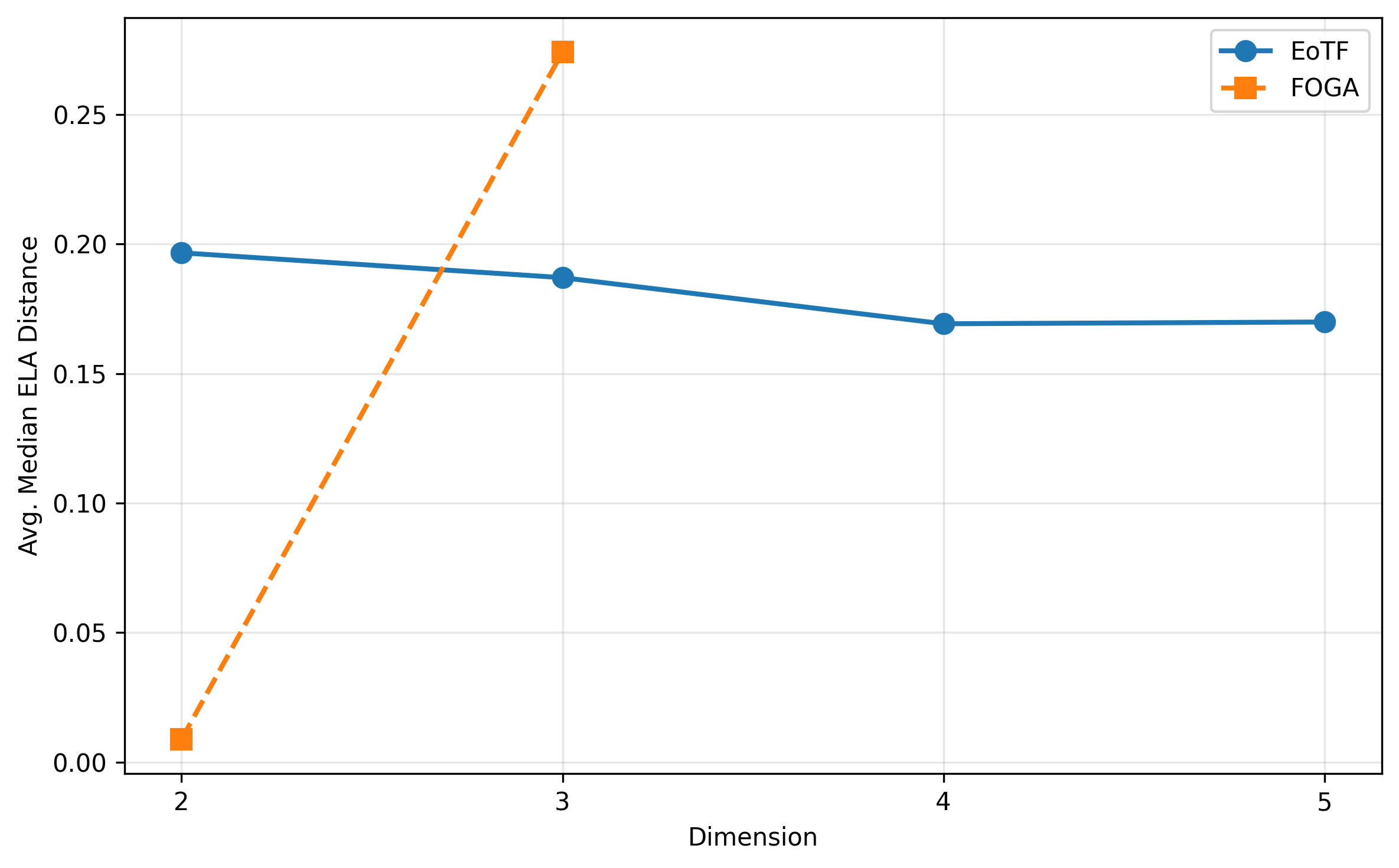}
\caption{Average median ELA distance across dimensions for EoTF and NNs on the BBOB suite.}
\label{fig:dimension_comparison}
\end{figure}

Figure~\ref{fig:dimension_comparison} illustrates the scalability of EoTF compared to the neural network-based approach proposed in \cite{foga_nn}. EoTF demonstrates remarkable stability in solution quality as dimensionality increases from 2D to 5D, with average median ELA distance remaining consistently between 0.17 and 0.20. This near-constant performance across dimensions suggests that EoTF does not suffer from the curse of dimensionality that typically affects landscape approximation methods.
In contrast, results for the neural network approach are available only for 2D and 3D functions due to prohibitive computational costs; as reported in \cite{foga_nn}, generating a single 2D point cloud requires up to 4 CPU hours, while a 3D point cloud demands nearly 3 CPU days. This computational barrier prevented evaluation in higher dimensions entirely.
Notably, while the neural network method achieves superior accuracy in two dimensions, its performance degrades substantially when extended to three dimensions, exceeding an average median ELA distance of 0.27. By comparison, EoTF achieves better solution quality in 5D (approximately 0.17) than the neural network approach attains in 3D. This demonstrates that although EoTF exhibits marginally lower accuracy in the lowest-dimensional case, it offers significantly more favorable scaling properties, a critical advantage for practical applications involving higher-dimensional optimization landscapes.

\section{Conclusions}

This paper introduced EoTF, a novel framework for the automated generation of benchmark functions guided by target ELA features. By leveraging LLMs within an evolutionary architecture, EoTF produces interpretable, self-contained Python functions that approximate prescribed landscape characteristics while circumventing the opacity inherent to NN-based approaches.

Our experimental evaluation addressed four research questions, yielding the following principal findings. First, EoTF demonstrates competitive performance against neural network baselines, with a pronounced advantage in higher-dimensional settings: while neural networks achieve superior accuracy in two dimensions, EoTF prevails on over 80\% of three-dimensional problems. Second, the method generalizes effectively to novel problem instances, as evidenced by consistent performance on the MA-BBOB suite, mitigating concerns regarding data contamination from memorized benchmark code. Third, more capable LLMs yield incremental improvements in generation quality, suggesting that continued advances in foundation models will further enhance the method's efficacy. Fourth, EoTF exhibits favorable scaling properties, maintaining stable solution quality from two to five dimensions, whereas the neural network baseline suffers from both degraded accuracy and prohibitive computational costs beyond three dimensions.

The interpretability of the generated artifacts constitutes a distinctive advantage of the proposed approach. Unlike neural network weights, the symbolic Python functions produced by EoTF permit direct inspection, theoretical analysis, and straightforward dissemination within the research community. Furthermore, the algorithm ranking experiments confirm that generated functions preserve the discriminative characteristics of the original benchmarks, validating their utility as analytical surrogates.

Several limitations warrant acknowledgment. NNs retain an accuracy advantage in the lowest-dimensional regime, suggesting that hybrid approaches may prove beneficial. Promising directions for future research include the systematic categorization of structural motifs in LLM-generated functions, the extension to constrained and multi-objective optimization landscapes, and the development of prompting strategies that explicitly target analytical properties such as specified optima locations or gradient characteristics. As LLM capabilities continue to advance and inference costs decline, we anticipate that the proposed methodology will become an increasingly practical tool for constructing the diverse, interpretable benchmarks required to rigorously evaluate and advance black-box optimization algorithms.

\printbibliography

@inproceedings{foga_nn,
author = {Prager, Raphael Patrick and Dietrich, Konstantin and Schneider, Lennart and Sch\"{a}permeier, Lennart and Bischl, Bernd and Kerschke, Pascal and Trautmann, Heike and Mersmann, Olaf},
title = {Neural Networks as Black-Box Benchmark Functions Optimized for Exploratory Landscape Features},
year = {2023},
isbn = {9798400702020},
publisher = {Association for Computing Machinery},
address = {New York, NY, USA},
url = {https://doi.org/10.1145/3594805.3607136},
doi = {10.1145/3594805.3607136},
booktitle = {Proceedings of the 17th ACM/SIGEVO Conference on Foundations of Genetic Algorithms},
pages = {129–139},
numpages = {11},
location = {Potsdam, Germany},
series = {FOGA '23}
}

@INPROCEEDINGS{gp_evolution,
  author={He, Yifan and Aranha, Claus},
  booktitle={2024 IEEE Congress on Evolutionary Computation (CEC)}, 
  title={Evolving Benchmark Functions to Compare Evolutionary Algorithms via Genetic Programming}, 
  year={2024},
  volume={},
  number={},
  pages={1-8},
  doi={10.1109/CEC60901.2024.10611801}}

@article{llamea,
author = {Stein, Niki van and B\"{a}ck, Thomas},
title = {LLaMEA: A Large Language Model Evolutionary Algorithm for Automatically Generating Metaheuristics},
year = {2024},
issue_date = {April 2025},
publisher = {IEEE Press},
volume = {29},
number = {2},
issn = {1089-778X},
url = {https://doi.org/10.1109/TEVC.2024.3497793},
doi = {10.1109/TEVC.2024.3497793},
journal = {Trans. Evol. Comp},
month = nov,
pages = {331–345},
numpages = {15}
}

@article{llm-sr,
  title={Llm-sr: Scientific equation discovery via programming with large language models},
  author={Shojaee, Parshin and Meidani, Kazem and Gupta, Shashank and Farimani, Amir Barati and Reddy, Chandan K},
  journal={arXiv preprint arXiv:2404.18400},
  year={2024}
}

@inproceedings{ela,
author = {Mersmann, Olaf and Bischl, Bernd and Trautmann, Heike and Preuss, Mike and Weihs, Claus and Rudolph, G\"{u}nter},
title = {Exploratory landscape analysis},
year = {2011},
isbn = {9781450305570},
publisher = {Association for Computing Machinery},
address = {New York, NY, USA},
url = {https://doi.org/10.1145/2001576.2001690},
doi = {10.1145/2001576.2001690},
booktitle = {Proceedings of the 13th Annual Conference on Genetic and Evolutionary Computation},
pages = {829–836},
numpages = {8},
location = {Dublin, Ireland},
series = {GECCO '11}
}

@inbook{malan_nn_benchmarking,
author = {Malan, Katherine Mary and Mu\~{n}oz, Mario Andr\'{e}s},
title = {Why We Should be Benchmarking Evolutionary Algorithms on Neural Network Training Tasks},
year = {2025},
isbn = {9798400714658},
publisher = {Association for Computing Machinery},
address = {New York, NY, USA},
url = {https://doi.org/10.1145/3712256.3726318},
booktitle = {Proceedings of the Genetic and Evolutionary Computation Conference},
pages = {30–38},
numpages = {9}
}

@inproceedings{hpo_x_ela,
author = {Schneider, Lennart and Sch\"{a}permeier, Lennart and Prager, Raphael Patrick and Bischl, Bernd and Trautmann, Heike and Kerschke, Pascal},
title = {HPO \texttimes{} ELA: Investigating Hyperparameter Optimization Landscapes by Means of Exploratory Landscape Analysis},
year = {2022},
isbn = {978-3-031-14713-5},
publisher = {Springer-Verlag},
address = {Berlin, Heidelberg},
url = {https://doi.org/10.1007/978-3-031-14714-2_40},
doi = {10.1007/978-3-031-14714-2_40},
booktitle = {Parallel Problem Solving from Nature – PPSN XVII: 17th International Conference, PPSN 2022, Dortmund, Germany, September 10–14, 2022, Proceedings, Part I},
pages = {575–589},
numpages = {15},
location = {Dortmund, Germany}
}

@article{COCO,
author = {Nikolaus Hansen and Anne Auger and Raymond Ros and Olaf Mersmann and Tea Tušar and Dimo Brockhoff},
title = {COCO: a platform for comparing continuous optimizers in a black-box setting},
journal = {Optimization Methods and Software},
volume = {36},
number = {1},
pages = {114--144},
year = {2021},
publisher = {Taylor \& Francis},
doi = {10.1080/10556788.2020.1808977},
URL = {https://doi.org/10.1080/10556788.2020.1808977},
eprint = {https://doi.org/10.1080/10556788.2020.1808977}
}

@article{funsearch,
  title={Mathematical discoveries from program search with large language models},
  author={Romera-Paredes, Bernardino and Barekatain, Mohammadamin and Novikov, Alexander and Balog, Matej and Kumar, M Pawan and Dupont, Emilien and Ruiz, Francisco JR and Ellenberg, Jordan S and Wang, Pengming and Fawzi, Omar and others},
  journal={Nature},
  volume={625},
  number={7995},
  pages={468--475},
  year={2024},
  publisher={Nature Publishing Group UK London}
}

@inproceedings{eoh,
author = {Liu, Fei and Tong, Xialiang and Yuan, Mingxuan and Lin, Xi and Luo, Fu and Wang, Zhenkun and Lu, Zhichao and Zhang, Qingfu},
title = {Evolution of heuristics: towards efficient automatic algorithm design using large language model},
year = {2024},
publisher = {JMLR.org},
booktitle = {Proceedings of the 41st International Conference on Machine Learning},
articleno = {1304},
numpages = {23},
location = {Vienna, Austria},
series = {ICML'24}
}

@inproceedings{detecting_funnel,
author = {Kerschke, Pascal and Preuss, Mike and Wessing, Simon and Trautmann, Heike},
title = {Detecting Funnel Structures by Means of Exploratory Landscape Analysis},
year = {2015},
isbn = {9781450334723},
publisher = {Association for Computing Machinery},
address = {New York, NY, USA},
url = {https://doi.org/10.1145/2739480.2754642},
doi = {10.1145/2739480.2754642},
booktitle = {Proceedings of the 2015 Annual Conference on Genetic and Evolutionary Computation},
pages = {265–272},
numpages = {8},
location = {Madrid, Spain},
series = {GECCO '15}
}

@inproceedings{fitness_distance,
author = {Jones, Terry and Forrest, Stephanie},
title = {Fitness Distance Correlation as a Measure of Problem Difficulty for Genetic Algorithms},
year = {1995},
isbn = {1558603700},
publisher = {Morgan Kaufmann Publishers Inc.},
address = {San Francisco, CA, USA},
booktitle = {Proceedings of the 6th International Conference on Genetic Algorithms},
pages = {184–192},
numpages = {9}
}

@article{pflacco,
    author = {Prager, Raphael Patrick and Trautmann, Heike},
    title = "{Pflacco: Feature-Based Landscape Analysis of Continuous and Constrained Optimization Problems in Python}",
    journal = {Evolutionary Computation},
    pages = {1-25},
    year = {2023},
    month = {07},
    issn = {1063-6560},
    doi = {10.1162/evco_a_00341},
    url = {https://doi.org/10.1162/evco\_a\_00341},
    eprint = {https://direct.mit.edu/evco/article-pdf/doi/10.1162/evco\_a\_00341/2148122/evco\_a\_00341.pdf},
}

@misc{li2025llameabolargelanguagemodel,
      title={LLaMEA-BO: A Large Language Model Evolutionary Algorithm for Automatically Generating Bayesian Optimization Algorithms}, 
      author={Wenhu Li and Niki van Stein and Thomas Bäck and Elena Raponi},
      year={2025},
      eprint={2505.21034},
      archivePrefix={arXiv},
      primaryClass={cs.LG},
      url={https://arxiv.org/abs/2505.21034}, 
}

@misc{alphaevolve,
      title={AlphaEvolve: A coding agent for scientific and algorithmic discovery}, 
      author={Alexander Novikov and Ngân Vũ and Marvin Eisenberger and Emilien Dupont and Po-Sen Huang and Adam Zsolt Wagner and Sergey Shirobokov and Borislav Kozlovskii and Francisco J. R. Ruiz and Abbas Mehrabian and M. Pawan Kumar and Abigail See and Swarat Chaudhuri and George Holland and Alex Davies and Sebastian Nowozin and Pushmeet Kohli and Matej Balog},
      year={2025},
      eprint={2506.13131},
      archivePrefix={arXiv},
      primaryClass={cs.AI},
      url={https://arxiv.org/abs/2506.13131}, 
}

@inproceedings{lmarena,
author = {Chiang, Wei-Lin and Zheng, Lianmin and Sheng, Ying and Angelopoulos, Anastasios N. and Li, Tianle and Li, Dacheng and Zhu, Banghua and Zhang, Hao and Jordan, Michael I. and Gonzalez, Joseph E. and Stoica, Ion},
title = {Chatbot arena: an open platform for evaluating LLMs by human preference},
year = {2024},
publisher = {JMLR.org},
booktitle = {Proceedings of the 41st International Conference on Machine Learning},
articleno = {331},
numpages = {30},
location = {Vienna, Austria},
series = {ICML'24}
}

@article{bench-MA-BBOB,
author = {Vermetten, Diederick and Ye, Furong and B\"{a}ck, Thomas and Doerr, Carola},
title = {MA-BBOB: A Problem Generator for Black-Box Optimization Using Affine Combinations and Shifts},
year = {2025},
issue_date = {March 2025},
publisher = {Association for Computing Machinery},
address = {New York, NY, USA},
volume = {5},
number = {1},
url = {https://doi.org/10.1145/3673908},
doi = {10.1145/3673908},
month = mar,
articleno = {5},
numpages = {19}
}

@misc{wang2025,
      title={Instance Generation for Meta-Black-Box Optimization through Latent Space Reverse Engineering}, 
      author={Chen Wang and Yue-Jiao Gong and Zhiguang Cao and Zeyuan Ma},
      year={2025},
      eprint={2509.15810},
      archivePrefix={arXiv},
      primaryClass={cs.LG},
      url={https://arxiv.org/abs/2509.15810}, 
}

@ARTICLE{munoz-space-filling,
  author={Muñoz, Mario A. and Smith-Miles, Kate},
  journal={Evolutionary Computation}, 
  title={Generating New Space-Filling Test Instances for Continuous Black-Box Optimization}, 
  year={2020},
  volume={28},
  number={3},
  pages={379-404},
  doi={10.1162/evco_a_00262}}

@ARTICLE{evolving-problems,
  author={Langdon, W. B. and Poli, Riccardo},
  journal={IEEE Transactions on Evolutionary Computation}, 
  title={Evolving Problems to Learn About Particle Swarm Optimizers and Other Search Algorithms}, 
  year={2007},
  volume={11},
  number={5},
  pages={561-578},
  doi={10.1109/TEVC.2006.886448}}

@unknown{cec,
author = {Liang, Jing and Suganthan, Ponnuthurai and Qu, B and Gong, D and Yue, Cai},
year = {2019},
month = {12},
pages = {},
title = {Problem Definitions and Evaluation Criteria for the CEC 2020 Special Session on Multimodal Multiobjective Optimization},
doi = {10.13140/RG.2.2.31746.02247}
}

@article{
symbolic-regression,
author = {Michael Schmidt  and Hod Lipson },
title = {Distilling Free-Form Natural Laws from Experimental Data},
journal = {Science},
volume = {324},
number = {5923},
pages = {81-85},
year = {2009},
doi = {10.1126/science.1165893},
URL = {https://www.science.org/doi/abs/10.1126/science.1165893},
eprint = {https://www.science.org/doi/pdf/10.1126/science.1165893}}

@article{autorank, doi = {10.21105/joss.02173}, url = {https://doi.org/10.21105/joss.02173}, year = {2020}, publisher = {The Open Journal}, volume = {5}, number = {48}, pages = {2173}, author = {Herbold, Steffen}, title = {Autorank: A Python package for automated ranking of classifiers}, journal = {Journal of Open Source Software} }

@misc{nevergrad,
    author = {J. Rapin and O. Teytaud},
    title = {{Nevergrad - A gradient-free optimization platform}},
    year = {2018},
    publisher = {GitHub},
    journal = {GitHub repository},
    howpublished = {\url{https://GitHub.com/FacebookResearch/Nevergrad}},
}

@inproceedings{Preuss_ELA_ASP,
  title={Algorithm selection based on exploratory landscape analysis and cost-sensitive learning},
  author={B. Bischl and Olaf Mersmann and Heike Trautmann and Mike Preuss},
  booktitle={Annual Conference on Genetic and Evolutionary Computation},
  year={2012},
  url={https://api.semanticscholar.org/CorpusID:5015211}
}

@ARTICLE{Tanabe_ELA_ASP,
  author={Tanabe, Ryoji},
  journal={IEEE Transactions on Evolutionary Computation}, 
  title={Benchmarking Feature-Based Algorithm Selection Systems for Black-Box Numerical Optimization}, 
  year={2022},
  volume={26},
  number={6},
  pages={1321-1335},
  doi={10.1109/TEVC.2022.3169770}
}

@article{team2024gemini,
  title={Gemini 1.5: Unlocking multimodal understanding across millions of tokens of context},
  author={Team, Gemini and Georgiev, Petko and Lei, Ving Ian and Burnell, Ryan and Bai, Libin and Gulati, Anmol and Tanzer, Garrett and Vincent, Damien and Pan, Zhufeng and Wang, Shibo and others},
  journal={arXiv preprint arXiv:2403.05530},
  year={2024}
}

@Article{         harris2020array,
 title         = {Array programming with {NumPy}},
 author        = {Charles R. Harris and K. Jarrod Millman and St{\'{e}}fan J.
                 van der Walt and Ralf Gommers and Pauli Virtanen and David
                 Cournapeau and Eric Wieser and Julian Taylor and Sebastian
                 Berg and Nathaniel J. Smith and Robert Kern and Matti Picus
                 and Stephan Hoyer and Marten H. van Kerkwijk and Matthew
                 Brett and Allan Haldane and Jaime Fern{\'{a}}ndez del
                 R{\'{i}}o and Mark Wiebe and Pearu Peterson and Pierre
                 G{\'{e}}rard-Marchant and Kevin Sheppard and Tyler Reddy and
                 Warren Weckesser and Hameer Abbasi and Christoph Gohlke and
                 Travis E. Oliphant},
 year          = {2020},
 month         = sep,
 journal       = {Nature},
 volume        = {585},
 number        = {7825},
 pages         = {357--362},
 doi           = {10.1038/s41586-020-2649-2},
 publisher     = {Springer Science and Business Media {LLC}},
 url           = {https://doi.org/10.1038/s41586-020-2649-2}
}

@misc{ael,
      title={Algorithm Evolution Using Large Language Model}, 
      author={Fei Liu and Xialiang Tong and Mingxuan Yuan and Qingfu Zhang},
      year={2023},
      eprint={2311.15249},
      archivePrefix={arXiv},
      primaryClass={cs.NE},
      url={https://arxiv.org/abs/2311.15249}, 
}

\appendix
\section{Prompt Templates} \label{app:prompts}

This appendix presents the complete prompt templates employed in our evolutionary framework for synthetic benchmark function generation. Following the EoH taxonomy~\cite{eoh}, we provide one initialization prompt (I1), two exploration prompts (E1, E2), and three mutation prompts (M1, M2, M3). Each prompt incorporates domain-specific instructions for Exploratory Landscape Analysis while preserving the original operator semantics.

All prompts share a common prefix that establishes the task context, specifies target ELA features, provides feature descriptions, and defines implementation requirements. For brevity, we present this shared component once in Prompt~\ref{pr:i1}, followed by the operator-specific instructions in Prompts~\ref{pr:e1}--\ref{pr:m3}.

\subsection{Dynamic Context Variables}

Each prompt template contains placeholders that are populated at runtime:

\begin{itemize}
    \item \texttt{\{ela\_features\}}: The target normalized ELA feature vector that the generated function should exhibit.
    \item \texttt{\{context\}}: Previously generated function(s) from the evolutionary process. The content varies by operator type. \textbf{Exploration prompts} (E1, E2) receive \emph{multiple} functions from the population history, encouraging the model to generate diverse alternatives or identify common structural patterns. \textbf{Mutation prompts} (M1, M2, M3) receive a \emph{single} parent function selected for modification, enabling focused refinement of promising candidates.
\end{itemize}

\begin{lstlisting}[style=promptstyle,caption={Prompt used to generate a synthetic benchmark function (I1) and a common part, shared preamble of the other prompts.},label={pr:i1}]
You are an expert in Exploratory Landscape Analysis (ELA), advanced optimization benchmarks, and high-dimensional function design.
Your task is to generate a single, synthetic benchmark function in Python for testing global optimization algorithms.

The primary goal is to create a function whose ELA features closely match the target values provided below.

Target Normalized ELA Features:
(These are the values the generated function's landscape should ideally exhibit)
{ela_features}

ELA Feature Descriptions:
- ela_meta.lin_simple.adj_r2: Adjusted R^2 of a linear model. High values suggest linearity.
- ela_meta.lin_w_interact.adj_r2: Adjusted R^2 of a linear model with pairwise interactions.
- ela_meta.quad_simple.adj_r2: Adjusted R^2 of a quadratic model without interactions.
- ela_meta.quad_w_interact.adj_r2: Adjusted R^2 of a full quadratic model.
- ela_distr.skewness: Skewness of the objective value distribution.
- nbc.nb_fitness.cor: Correlation between fitness and nearest-better connectivity.
- nbc.nn_nb.sd_ratio: Ratio of standard deviations (nearest neighbor distance / nearest-better distance).
- fitness_distance.fitness_std: Standard deviation of objective values.

Implementation Requirements:
1. Language & Libraries: Implement the function in Python, using only NumPy for mathematical operations.
2. Function Signature:
   ```python
   def problem(x: np.ndarray) -> float:
       # Docstring goes here
       pass
   ```
   Your code MUST BE included in a markdown code block.
3. Input: x is a 1D NumPy array of shape (N,).
4. Domain: The function should be designed considering the domain [-5, 5]^N. Ensure operations are valid within this domain.
5. Docstring: Include a concise docstring explaining the mathematical structure of the function. If possible, include the formula. Be specific about the components used.
6. Self-Contained Code: The final output block should only contain the necessary import (import numpy as np) and the function definition.
7. The function must be deterministic: Do not use np.random or any stochastic elements.
\end{lstlisting}

\begin{lstlisting}[style=promptstyle,caption={Exploration Prompt (E1): Divergent Generation},label={pr:e1}]
{I1}

History:
You already generated these functions:
{context}

Instructions:
Please help me create a new function that has a totally different form from the given ones.
\end{lstlisting}

\begin{lstlisting}[style=promptstyle,caption={Exploration Prompt (E2): Backbone Extraction},label={pr:e2}]
{I1}

History:
You already generated these functions:
{context}

Instructions:
Please help me create a new function that has a totally different form from the given ones but can be motivated from them. Firstly, identify the common backbone idea in the provided functions. Secondly, based on the backbone idea create a new solution.
\end{lstlisting}

\newpage

\begin{lstlisting}[style=promptstyle,caption={Mutation Prompt (M1): Structural Modification},label={pr:m1}]
{I1}

Generated Function:
You already generated this function:
{context}

Instructions:
Please assist me in creating a new function that has a different form but can be a modified version of the function provided.
\end{lstlisting}

\begin{lstlisting}[style=promptstyle,caption={Mutation Prompt (M2): Parameter Refinement},label={pr:m2}]
{I1}

Generated Function:
You already generated this function:
{context}

Instructions:
Please identify the main parameters of the generated function and assist me in creating a new version of the function with improved parameter settings.
\end{lstlisting}

\begin{lstlisting}[style=promptstyle,caption={Mutation Prompt (M3): Simplification},label={pr:m3}]
{I1}

Generated Function:
You already generated this function:
{context}

Instructions:
First, you need to identify the main components in the function above. Next, analyze whether any of these components can be overfit to the specific sample of points used to calculate ELA features. Then, based on your analysis, simplify the components to enhance the generalization to other samples.
\end{lstlisting}

\end{document}